\documentclass{article}

\usepackage[numbers]{natbib}
\usepackage{arxiv}

\usepackage[utf8]{inputenc} 
\usepackage[T1]{fontenc}    
\usepackage{hyperref}       
\usepackage{url}            
\usepackage{booktabs}       
\usepackage{amsfonts}       
\usepackage{nicefrac}       
\usepackage{microtype}      
\usepackage{lipsum}		
\usepackage{graphicx}
\usepackage{natbib}
\usepackage{doi}
\usepackage{caption}
\usepackage{ragged2e}
\usepackage{hyperref} 
 \hypersetup{ 
     colorlinks=true, 
     linkcolor=blue, 
     filecolor=blue, 
     citecolor = black,       
     urlcolor=blue, 
     } 
\usepackage{hyperref}
\usepackage{array}
\usepackage{booktabs}
\usepackage{longtable}
\usepackage{geometry}
\usepackage{enumitem}
\usepackage{textcomp}
\usepackage{listings}

\usepackage{newtxmath}
\usepackage{tikz}
\usetikzlibrary{trees}
\usepackage{hyperref} 

\lstdefinestyle{smalltypewriter}{
  basicstyle=\footnotesize\ttfamily, 
  breaklines=true,                     
  columns=fullflexible,                
  keepspaces=true,                     
  xleftmargin=0pt,                     
  xrightmargin=0pt,                    
  frame=single,                        
}

\title{FAIR Enough: Develop and Assess a FAIR-Compliant Dataset for Large Language Model Training? }

\author{
  Shaina Raza\thanks{Corresponding author. Email: shaina.raza@vectorinstitute.ai} \\
  Vector Institute for Artificial Intelligence\\
  Toronto, ON, Canada \\
  \texttt{shaina.raza@vectorinstitute.ai} \And
   Shardul Ghuge \\
  Vector Institute for Artificial Intelligence\\
  Toronto, ON, Canada \\
  \texttt{shardul.ghuge@mail.utoronto.ca} \And
  Chen Ding \\
  Toronto Metropolitan University\\
  Toronto, ON, Canada \\
  \texttt{chen.ding@torontomu.ca}  \And
  Elham Dolatabadi\\
  Vector Institute for Artificial Intelligence\\
  York University
  Toronto, ON, Canada \\
  \texttt{edolatab@yorku.ca} \And
  Deval Pandya \\
  Vector Institute for Artificial Intelligence\\
  Toronto, ON, Canada \\
  \texttt{deval.pandya@vectorinstitute.ai} 
}

\renewcommand{\shorttitle}{\textit{arXiv} Template}

\hypersetup{
pdftitle={A template for the arxiv style},
pdfsubject={q-bio.NC, q-bio.QM},
pdfauthor={David S.~Hippocampus, Elias D.~Striatum},
pdfkeywords={First keyword, Second keyword, More},
}

\begin{document}
\maketitle

\begin{abstract}
The rapid evolution of Large Language Models (LLMs) highlights the necessity for ethical considerations and data integrity in AI development, particularly emphasizing the role of FAIR (Findable, Accessible, Interoperable, Reusable) data principles. While these principles are crucial for ethical data stewardship, their specific application in the context of LLM training data remains an under-explored area. This research gap is the focus of our study, which begins with an examination of existing literature to underline the importance of FAIR principles in managing data for LLM training. Building upon this, we propose a novel framework designed to integrate FAIR principles into the LLM development lifecycle. A contribution of our work is the development of a comprehensive checklist intended to guide researchers and developers in applying FAIR data principles consistently across the model development process. The utility and effectiveness of our framework are validated through a case study on creating a FAIR-compliant dataset aimed at detecting and mitigating biases in LLMs. We present this framework to the community as a tool to foster the creation of technologically advanced, ethically grounded, and socially responsible AI models.
\end{abstract}




\maketitle

\section{Introduction}
In the era of artificial intelligence (AI), Language Models (LMs) play an important role in advancing diverse AI applications \cite{jiang2020can}. From virtual assistants to content generation, LMs have become indispensable in  shaping both the future trajectory of academic research and in their widespread practical applications. The impact of this transformation is further amplified by the evolution of Large Language Models (LLMs) \cite{zhao2023survey}, such as OpenAI's GPT series, Gemini, LLaMa, and Falcon. As of January 2024, LLM development has collected \$18.2 billion in funding and \$2.1 billion in revenue \cite{trendfeedr_llm_trends}.

The rapid success of these LLMs highlight the importance of diverse data for broadening their applicability across different domains. This idea is also well-supported in research \cite{bender2021dangers,wang2023aligning}, which affirms the importance of high-quality data in training these models. However, the training of these big models on data from diverse sources highlights the complex ethical and responsible data practice challenges in their implementation  \cite{chang2023survey,wang2023aligning}.

The FAIR data principles, which stand for Findable, Accessible, Interoperable, and Reusable \cite{dunning1970fair,boeckhout2018fair}, were initially established to improve the stewardship of scientific data. These principles are can be used for any model development life-cycle \cite{wise2019implementation,chen2022implementing} and have become increasingly recognized in responsible AI development \cite{deshpande2022responsible}. Recently, the relevance of FAIR principles has been particularly highlighted in generative AI due to ethical challenges such as bias, privacy concerns, and the potential misuse of AI-generated content \cite{openai_ethics, partescano2021data}. This highlights the growing importance of ensuring data used in building these LLMs is findable, accessible, interoperable, and reusable, adhering to ethical standards. 'LLMs' here includes both 'LMs' and 'LLMs', with 'LLMs' representing more advanced versions.

Seminal works in data science and management \cite{dunning1970fair, wilkinson2016fair, hasnain2018assessing, jacobsen2020fair, FairPayNLP2021} have explored FAIR principles in various domains. Their application in the training and development of LLMs, however, is a developing area of research. While recent advances in LLM studies \cite{chang2023survey,wang2023aligning} have focused on aligning LLMs with ethical standards including human values, the direct incorporation of FAIR principles in LLM training is less explicit. Challenges specific to LLMs, such as addressing data biases, toxicity, stereotypes and the need for model explainability and interpretability \cite{singh2023augmenting}, further highlight the need for ethical and balanced training approaches. This emphasis on data ethics is in line with broader trends in AI ethics and data science \cite{raji2021ai, jobin2019global} stress the necessity of integrating these principles into the LLM development.

In this work, we shed some light on FAIR data principles and propose a model development lifecycle for LLM training that incorporates FAIR principles at each stage. One of the contributions of this work is the development of a \textit{FAIR-compliant dataset}, designed to include a wide array of narratives from diverse sources.  To emphasize the significance of FAIR data principles in the context of preparing this dataset for LLM training, it is crucial to understand that while this process may not directly equate to the complete ``FAIRification" of LLMs, it represents a critical and foundational step towards it. The application of FAIR principles ensures that the data feeding into LLMs is of high quality and organized in a way that maximizes its utility, thereby enhancing the model's performance and reliability.

The central contribution of our work, however, extends beyond simple data collection and preparation for LLM training. We have placed a strong emphasis on rigorously aligning the LLM training dataset with FAIR principles. We acknowledge that adherence to FAIR principles (even at the strict-most level) may not equate to absolute ethical compliance, however, it represents a crucial step in that direction. Our research establishes a foundational framework for further advanced studies in this field. The primary contributions of our work are:

\subsection{Contributions}
\begin{enumerate}
    \item An exploration of FAIR data principles in general AI research, including the provision of a comprehensive checklist for researchers and developers.
    \item Introduction of an innovative framework that integrates FAIR data principles throughout the LLM training lifecycle, ensuring ethical and effective application in various AI contexts.
    \item Demonstration of the practical benefits of a FAIR-compliant dataset through a case study. This study specifically focuses on identifying and mitigating biases prior to training LLMs.  The topic of bias is broad-ranging, our case study specifically focuses on addressing linguistic biases targeting protected groups.

\end{enumerate}

The data used for pre-training LLMs is predominantly unstructured. However, as demonstrated in this case study, if it is accurately labeled and formatted—interoperability being one such key principle—it can then be leveraged for fine-tuning a wide array of downstream tasks. Furthermore, adherence to FAIR principles not only ensures data is handled correctly but also significantly boosts the credibility of the models.

\subsection{Literature Selection Criteria }
This study investigates articles published within the past five years, focusing on the rise of generative AI models like those in the GPT series. The study specifically targets English-language articles from leading AI and ML journals and conferences.
We explored several databases and proceedings from high-quality journals from Elsevier, Springer, Nature portfolio, IEEE Transactions, and key conferences like NeurIPS, ICML, ACL and such. Recognizing the fast-paced evolution of LLM research, we included relevant preprints and seminal works for a comprehensive perspective. This search resulted in 135 papers, out of which we carefully chose about 75 that directly pertained to FAIR data principles in the context of training models. We noted a scarcity of work combining LLMs with FAIR data principles, a gap that our study aims to address. 

The search query for this study is :

\begin{lstlisting}[style=smalltypewriter]
("Findability" OR "data discovery" OR "metadata standards" OR "persistent identifiers") AND ("Accessibility" OR "data access" OR "data sharing policies" OR "authentication and authorization mechanisms") AND ("Interoperability" OR "data integration" OR "standardized data formats" OR "cross-domain data exchange") AND ("Reusability" OR "data documentation" OR "data quality assurance" OR "long-term data preservation") AND ("large language models" 
OR "LLMs" OR "AI models" OR "machine learning models") AND ("training data management" OR "ethical data sourcing" OR "bias in AI datasets" OR "responsible data use in AI") AND ("ethical considerations in AI" OR "AI ethics" OR "responsible AI" OR "ethical AI development") PUBLISHED FROM 2018 TO 2023 IN English IN Journals and Conferences (e.g., JAIR, JMLR, IEEE Transactions, NeurIPS, ICLR, ACL, Springer, Nature portfolio, Elsevier).
\end{lstlisting}

\subsection{Comparative Analysis of Related Work}
We have conducted a meta-review of key articles on FAIR data principles, categorizing them by domain and their relevance to LLMs. We also highlight the difference of our work compared to the previous works. The findings are summarized in Table \ref{table:fair_data_principles}.

\begin{table}[h]
\small
\centering
\caption{Summary of Review Papers on FAIR Data Principles}
\begin{tabular}{|p{1.5cm}|p{5.5cm}|p{2cm}|p{2.5cm}|}
\hline
 \textbf{Reference} & \textbf{Summary} & \textbf{Domain Focus} & \textbf{LLM Relevance} \\ \hline
  \cite{alvarez2023desiderata} & Discusses data governance for FAIR principles in health data management.& Health Data & Not Applicable \\ \hline
  \cite{inau2021initiatives} & Reviews initiatives for applying FAIR principles in managing health data. & Health Data & Not Applicable \\ \hline
  \cite{sadeh2023opportunities} & Examines FAIR data practices' role in global mental health research. & Mental Health Research & Not Applicable \\ \hline
  \cite{stanciu2023data} & Analyzes FAIR principles in healthcare data, emphasizing cybersecurity. & Health Data & Partially (InstructGPT) \\ \hline
  \cite{raycheva2023challenges} & Addresses challenges in FAIR application and legal aspects in European rare disease databases for ML technologies. & Health Data & Not Applicable \\ \hline
  \cite{vesteghem2020implementing} & Elaborates on FAIR principles in precision oncology research.& Health Data & Not Applicable \\ \hline
  \cite{jacobsen2020fair} & Offers a broad perspective on FAIR principles' implementation across various fields.& Multi-Domain & Not Applicable \\ \hline
  Our Study & Evaluates the application of FAIR data principles in the creation and training of datasets for LLMs. This study encompasses a range of models including the BERT and GPT families, as well as the LLaMa-2-7b model, highlighting diverse approaches in LLM development.& News Data & Yes (BERT/   , GPT families, LLaMa-2-7b)\\ \hline
\end{tabular}

\label{table:fair_data_principles}
\end{table}
Table \ref{table:fair_data_principles} highlights a focus on health data in existing research on FAIR principles. One study incorporates a LLM application (GPT-based model). Our work, however, stands out for its application of FAIR data principles to a range of LLMs, including pretrained LMs, GPT-series, and Llama2 models. This approach broadens the scope and generalizability of our research.

\section{FAIR Data Principles: Theoretical Background and Significance}

\begin{figure}[h]
    \centering
    \includegraphics[width=0.8\linewidth]{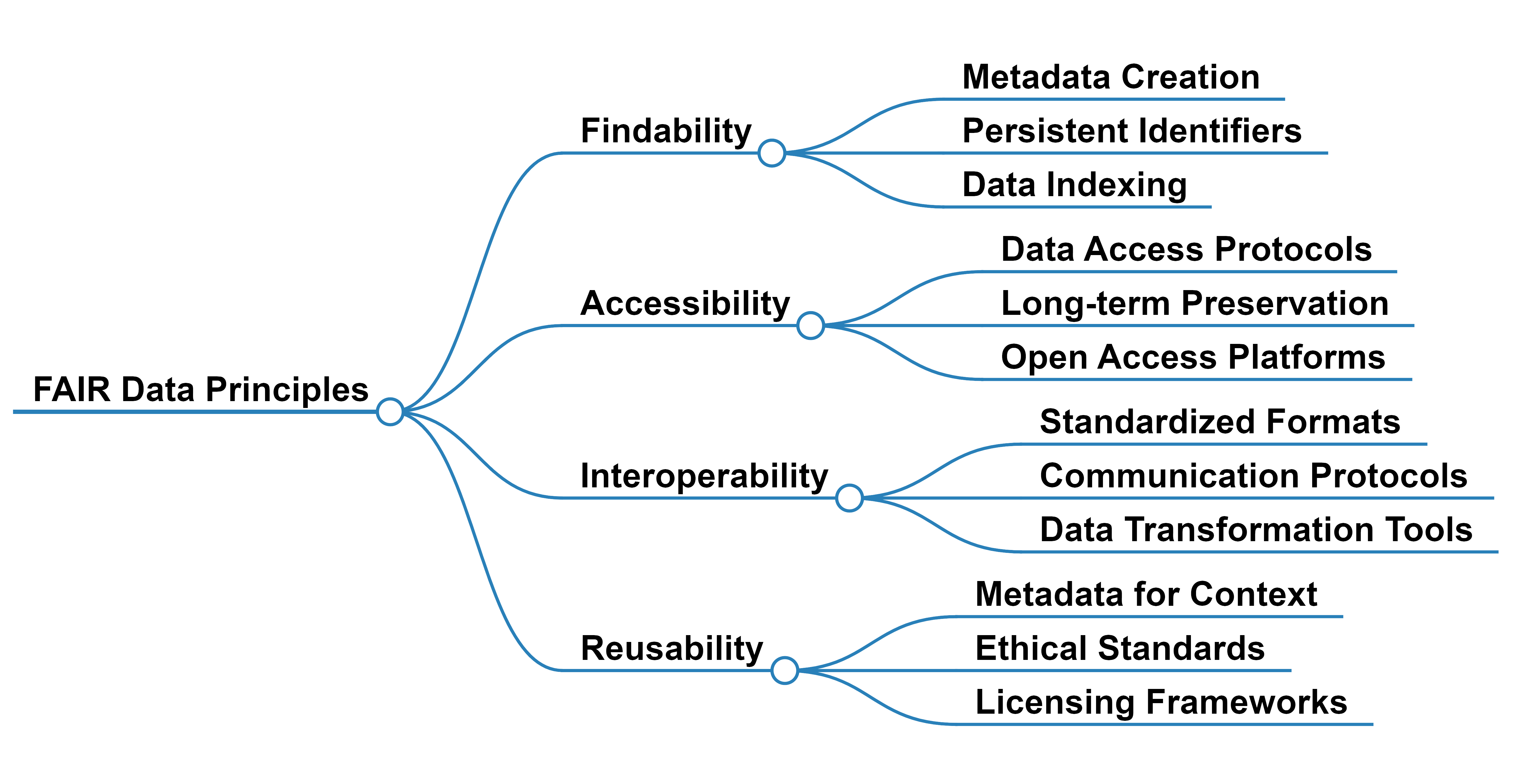}
    \caption{FAIR Data Principles: Key Aspects of Findability, Accessibility, Interoperability, and Reusability in Data Management. }
    \label{fig:mindmap1}
\end{figure}

The FAIR data principles significantly influence data management practices across various fields. These include physics \cite{FAIRPrinciplesHighEnergy2022}, health \cite{inau2023initiatives}, environmental science \cite{partescano2021data}, pharmaceuticals \cite{wise2019implementation}, chemistry \cite{jeliazkova2023fair}, computer science \cite{axton2016fair}, research and development \cite{wilcox2018supporting}, and clinical studies \cite{vesteghem2020implementing}. These principles ensure data is well-organized, accessible, and reusable in different research domains and applications. In AI and Natural Language Processing (NLP), the implementation of interdisciplinary strategies \cite{FAIRforAI2022} and structured approaches \cite{GoalOrientedFAIR2023} is crucial to uphold ethical data practices \cite{FairPayNLP2021} and ensure fairness in AI models \cite{AIFairness2022}. Given the constraints of space, we present a concise overview of each FAIR principle, as depicted in Figure \ref{fig:mindmap1}. Additionally, a compliance checklist is provided in Table \ref{table:fair_comprehensive_checklist}.For hyperlinks to the technologies and APIs referenced in the table, refer to Appendix \ref{App: acronyms}.

\subsection{Findability}

The principle of Findability is important for ensuring that data and resources are not only discoverable but also readily accessible \cite{findlay2020ecosystem}. It involves the establishment of detailed metadata, the implementation of persistent identifiers, and the promotion of effective data indexing and search strategies. This principle significantly enhances the discoverability of data by both humans and machines, facilitating a smoother research process. Recent contributions to the enhancement of data findability within the FAIR framework include research on strategies for the FAIRification \cite{GoalOrientedFAIR2023} of data to improve its findability, and studies \cite{santos2023towards} investigating models for the FAIR digital object framework with a focus on improving the findability of digital objects. Discussions \cite{FAIRforAI2022} on the application of FAIR principles specifically within AI datasets highlight the growing recognition of these principles' importance. Additionally, considerations on the trustworthiness of data repositories underline findability's critical role in ensuring data quality and reliability in these environments \cite{Gotz2023Fair}.

\subsection{Accessibility}
Accessibility, as articulated by the FAIR data principles, emphasizes the straightforwardness of obtaining and utilizing data upon its discovery \cite{boeckhout2018fair}. This aspect is integral to the FAIR framework and encompasses strategies for long-term data preservation, the establishment of ethical access protocols, and the assurance of data retrievability and usability across time. Key discussions  \cite{wang2023fair} on the role of trust in data repositories, an essential component of accessibility, offering a detailed exploration of accessibility within the context of the FAIR principles. The application of FAIR principles to research software is also discussed in the literature \cite{lamprecht2020towards} with a specific lens on enhancing accessibility. The support of the Fedora platform for FAIR data principles, especially in terms of accessibility, is examined in a related work \cite{wilcox2018supporting}. Furthermore, the challenges and opportunities encountered in the implementation of FAIR principles, including accessibility, within the Brazilian data science landscape is also discussed \cite{sales2020go}. Collectively, these contributions highlight the critical importance of embedding accessibility considerations into research software, platform support, and addressing implementation hurdles.

\subsection{Interoperability}
Interoperability, a core aspect of the FAIR data principles, denotes the capability of different data systems to operate cohesively \cite{da2016fair}. This principle necessitates the adoption of standardized data formats and protocols to facilitate straightforward data exchange and integration across heterogeneous systems. Discussions on the Immune Epitope Database (IEDB) emphasize its commitment to interoperability through the adoption of FAIR principles \cite{vita2018fair}. The development and support of interoperability in ontology creation, as facilitated by the eXtensible ontology development (XOD) principles and tools, are detailed in \cite{he2018extensible}. Furthermore, the examination of challenges and strategies related to the implementation of FAIR data principles, with a particular focus on interoperability in data science, is undertaken in \cite{sales2020go}. Additionally, a framework along with metrics for assessing the FAIRness of data, emphasizing interoperability, are delineated in \cite{wilkinson2018design}. The objectives and efforts of the GO FAIR initiative, aimed at promoting the widespread adoption of FAIR principles, including interoperability, are elaborated in \cite{schultes2018ready}. Collectively, these contributions highlight the role of interoperability in promoting responsible and efficient data management across diverse scientific and technological domains.

\subsection{Reusability}
Reusability, as one of the foundational aspects of the FAIR data principles, highlights the necessity for data to be stored and documented in a manner that facilitates future retrieval and reuse \cite{anguswamy2012study}. This principle is supported by the creation of comprehensive metadata, consideration of legal and ethical frameworks, and the assessment of potential societal impacts. Research focusing on a structured methodology for planning the FAIRification of data, particularly with reusability in mind, is presented in \cite{GoalOrientedFAIR2023}. Furthermore, the development of a model for digital objects adhering to FAIR principles that prioritizes reusability is detailed in \cite{santos2023towards}. The exploration of FAIR principles for dataset reusability, is disussed in \cite{FAIRforAI2022}. Additionally, the workflows that prioritize reusability from the outset is argued in \cite{wolf2021reusability}. 
An exploration of FAIR principles, including detailed insights on reusability, is presented in \cite{jacobsen2020fair}. Collectively, these contributions advocate for a data management approach that not only preserves data for future use but also ensures it remains ethically and effectively employable for training.

\begin{small}
\centering
\begin{longtable}{|p{2cm}|p{2.3cm}|p{4cm}|p{3.5cm}|}
\caption{Comprehensive Checklist for Ensuring FAIR Principles Compliance, Along with Associated Tools and Practices.} \label{table:fair_comprehensive_checklist} \\
\hline
\textbf{Principle} & \textbf{Features} & \textbf{Compliance list} & \textbf{Tools and Practices} \\
\hline
\endfirsthead

\multicolumn{4}{c}%
{{\bfseries Table \thetable\ Continued from previous page}} \\
\hline
\textbf{Principle} & \textbf{Features} & \textbf{Compliance list} & \textbf{Tools and Practices} \\
\hline
\endhead

\hline
\endfoot

Findability & 
   F1: Rich and descriptive metadata \newline
   F2: Standardized data indexing \newline
   F3: Documented data sources \newline
   F4: Advanced search functionalities \newline
   F5: Persistent identifiers like DOIs
& 
   F1: Metadata includes titles, authors, abstracts, keywords, and affiliations. \newline
   F2: Use of standardized taxonomy and ontology for indexing. \newline
   F3: Clear documentation of data sources and collection methodologies. \newline
   F4: Implementation of advanced search tools and interfaces. \newline
   F5: Assignment of persistent identifiers such as DOIs.
& 
   Metadata Management: \href{https://atlas.apache.org}{Apache Atlas}, \href{https://www.collibra.com/}{Collibra}, \href{https://orcid.org/}{ORCID} \newline
   Persistent Identifiers: \href{https://www.crossref.org/}{CrossRef} for DOIs \newline
   Data Indexing: \href{https://www.elastic.co/elasticsearch/}{Elasticsearch}, \href{https://solr.apache.org/}{Apache Solr}, \href{https://duraspace.org/dspace/}{DSpace} \newline
   Search Interfaces: \href{https://www.algolia.com/}{Algolia}, \href{https://lucene.apache.org/}{Apache Lucene} \newline
   Data Repositories: \href{https://www.ncbi.nlm.nih.gov/}{NCBI}, \href{https://www.re3data.org/}{RE3data}, \href{https://ckan.org/}{CKAN}, \href{https://dataverse.org/}{Dataverse}, \href{https://zenodo.org/}{Zenodo}, \href{https://figshare.com/}{Figshare}, \href{https://www.eprints.org/}{EPrints}, \href{https://www.researchgate.net/}{ResearchGate}, \href{https://www.academia.edu/}{Academia.edu}
\\
\hline
Accessibility & 
   A1: Clear data access protocols \newline
   A2: Long-term data preservation \newline
   A3: Open access platforms \newline
   A4: Standardized APIs for data access \newline
   A5: Data availability in accessible formats
& 
   A1: Detailed access instructions and authentication processes. \newline
   A2: Use of reliable digital preservation services like CLOCKSS or Portico. \newline
   A3: Data deposited in open repositories such as Figshare or Zenodo. \newline
   A4: APIs conform to standards such as OpenAPI for ease of use. \newline
   A5: Data provided in multiple formats (e.g., CSV, JSON, XML) to ensure usability.
& 
  API Tools: \href{https://www.openapis.org/}{OpenAPI}, \href{https://graphql.org/}{GraphQL}, RESTful interfaces \newline
   Data Preservation: \href{https://www.archivematica.org/en/}{Archivematica}, \href{https://www.lockss.org/}{LOCKSS}, \href{https://zenodo.org/}{Zenodo}, \href{https://figshare.com/}{Figshare} \newline
   Cloud Storage: \href{https://aws.amazon.com/s3/}{Amazon S3}, \href{https://cloud.google.com/storage}{Google Cloud Storage}, \href{https://azure.microsoft.com/en-us/services/storage/}{Microsoft Azure} \newline
   Ethical Access: \href{https://www.onetrust.com/}{OneTrust}, \href{https://www.trustarc.com/}{TrustArc}
\\
\hline
Interoperability & 
   I1: Standard data formats \newline
   I2: Common communication protocols \newline
   I3: Data exchange standards \newline
   I4: Tools for data transformation and mapping
& 
   I1: Data conforms to community-recognized standards (e.g., MIAME, Ecological Metadata Language). \newline
   I2: Support for protocols such as OAI-PMH for metadata harvesting. \newline
   I3: Use of frameworks like Schema.org for structured data. \newline
   I4: Availability of services like XSLT or OpenRefine for data conversion and mapping.
& 
  Data Formats: JSON, XML, CSV, DICOM, GenBank \newline
   Protocols: HTTP, SOAP, \href{https://restfulapi.net/}{REST}, \href{https://grpc.io/}{gRPC} \newline
   Data Standards: RDF, \href{https://www.hl7.org/fhir/}{HL7 FHIR}, ISO/IEC standards \newline
   Transformation Tools: \href{https://www.w3schools.com/xml/xsl_intro.asp}{XSLT}, \href{https://www.talend.com/}{Talend}, \href{https://www.informatica.com/}{Informatica}, \href{https://nifi.apache.org/}{Apache NiFi} \newline
   Ontology Systems: OWL, \href{https://www.w3.org/TR/sparql11-query/}{SPARQL}, \href{https://xod.io/}{XOD} \newline
   Specialized Databases: \href{https://www.iedb.org/}{IEDB}
\\
\hline
Reusability & 
   R1: Detailed metadata for context \newline
   R2: Data curated for future use \newline
   R3: Adherence to ethical standards \newline
   R4: Licensing frameworks \newline
   R5: Consideration of societal impacts
& 
   R1: Comprehensive metadata including experimental conditions, methodologies, and provenance. \newline
   R2: Curating data with clear versioning and update records. \newline
   R3: Compliance with GDPR and other privacy regulations. \newline
   R4: Use of licenses like Creative Commons to clarify user rights. \newline
   R5: Assessment and documentation of data impact on society and potential biases.
& 
Metadata Standards: \href{https://www.dublincore.org/}{Dublin Core}, \href{https://www.datacite.org/}{DataCite}, \href{https://schema.org/}{schema.org} \newline
   Data Curation: \href{https://ckan.org/}{CKAN}, \href{https://duraspace.org/dspace/}{DSpace}, \href{https://omeka.org/}{Omeka} \newline
   Ethical Frameworks: \href{https://www.responsible.ai/}{Responsible AI}, \href{https://openai.com/ethics/}{OpenAI Ethics Guidelines}, \href{https://ai-4-all.org/}{AI4ALL} \newline
   Licensing Tools: \href{https://creativecommons.org/}{Creative Commons} \newline
  Data Provenance tools: Provenance Tracking \href{https://www.w3.org/TR/prov-dm/}{PROV-DM}\\
\hline
\end{longtable}
\end{small}

\section{Data Management Challenges in Large Language Models}

\noindent The evolution of LLMs introduces a complex spectrum of data management challenges, necessitating advanced strategies for efficient data organization and accessibility due to the engagement with extensive and intricate datasets \cite{zhao2023survey}. The importance of high-quality data cannot be overstated; it is highly important to use representative, well-curated datasets to develop models that are ethically sound and have broad applicability \cite{bender2021dangers}. Addressing privacy and ethical concerns is essential, requiring rigorous data governance to adhere to ethical norms and protect individual privacy \cite{wang2023aligning}. Furthermore, the precision of data annotation and labeling is important to ensure model reliability, calling for standardized and transparent practices \cite{raza2023constructing, monarch2021human}. Balancing data accessibility with the protection of proprietary information is crucial, a task that must be aligned with legal and ethical standards \cite{jobin2019global}. Moreover, compliance with data protection laws is critical for legal conformity and upholding the ethical integrity of AI technologies, which is fundamental to sustaining public trust \cite{raji2021ai}. A structured overview of these data management challenges and requirements in LLM development is provided in Figure \ref{mindmap2}, with a summary given below.

\begin{figure}
    \centering
    \includegraphics[width=0.85\linewidth]{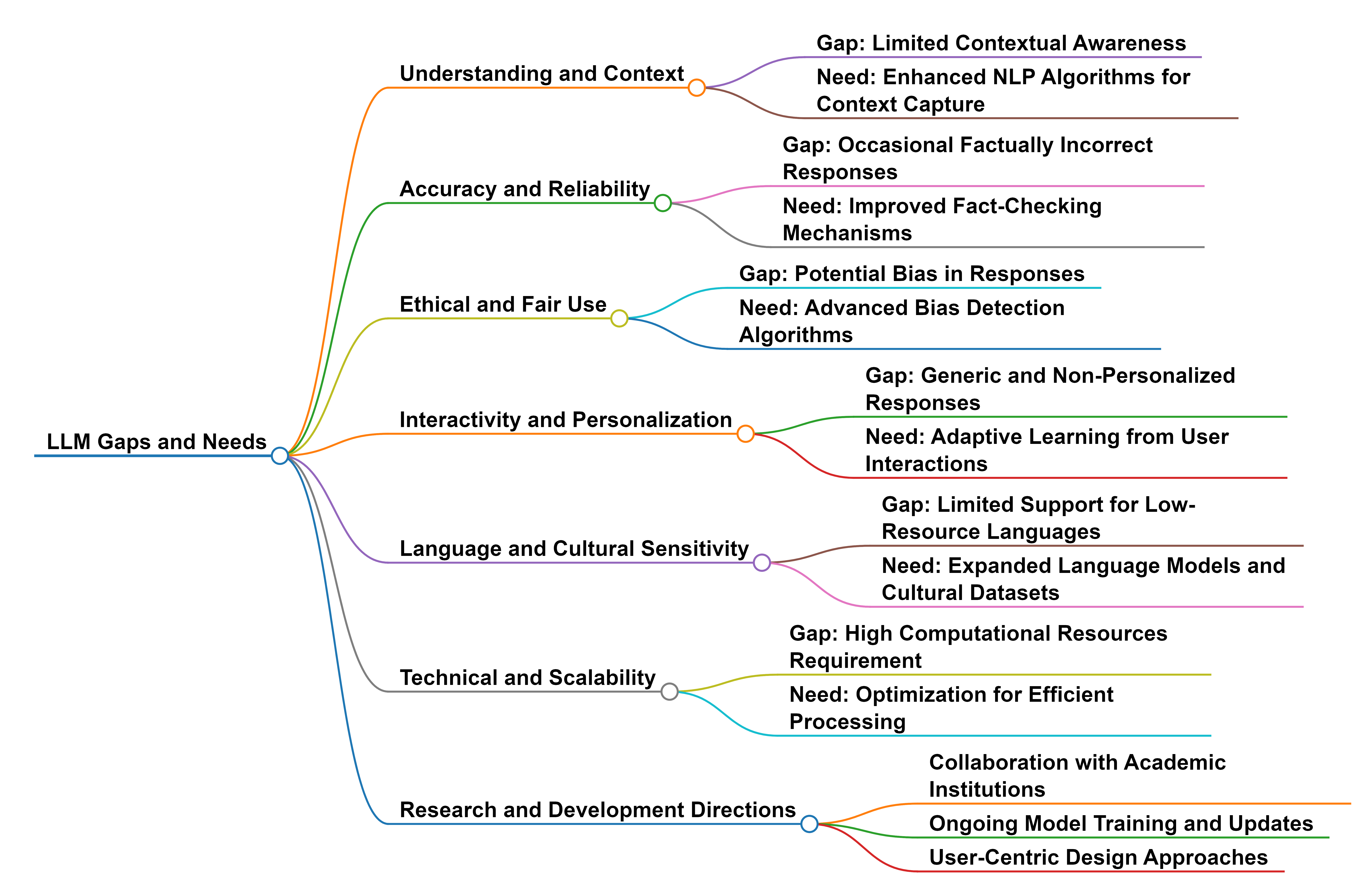}
    \caption{Data Management Challenges in Large Language Models}
    \label{mindmap2}
\end{figure}

\textit{Understanding the Context}: 
Improving algorithms and reasoning is essential for enhancing contextual awareness in LLMs \cite{xi2023rise}. The capability of these models to understand and interpret complex contexts significantly contributes to their practical utility. Developing algorithms that grasp linguistic nuances, cultural contexts, and the implications of language use is increasingly necessary.

\textit{Accuracy and Reliability}: 
LLMs sometimes producing incorrect information highlights the need for effective fact-checking mechanisms \cite{ji2023survey}. To increase the reliability of these models, it is vital to implement verification algorithms and data validation techniques that ensure their outputs are accurate and trustworthy.

\textit{Ethical and Fair Use}: 
Mitigating biases in LLM outputs is a significant concern \cite{deshpande2022responsible,wang2023aligning,raza2022dbias}. Developing algorithms for detecting and correcting biases in training data and model outputs is essential for fair use \cite{raza2023fairness}. Efforts must focus on incorporating diverse perspectives to ensure ethical use and fairness in model responses.

\textit{Interactivity and Personalization}: 
The general nature of responses from current LLMs indicates the importance of developing adaptive learning algorithms \cite{he2023large}. Such algorithms should learn from user interactions, preferences, and feedback to provide personalized responses \cite{porsdam2023autogen}. The aim is to create models that tailor their responses to individual user needs, improving personalization and user experience.

\textit{Language and Cultural Sensitivity}: 
Addressing the underrepresentation of low-resource languages is crucial for increasing the inclusivity of LLMs \cite{chang2023survey}. It is important to enhance language model diversity and cultural dataset representation \cite{ranathunga2023neural}. Expanding datasets to include more languages and ensuring models respect cultural contexts are key steps toward global inclusivity.

\textit{Efficiency and Scalability}: 
The significant computational demands of LLM development present a notable challenge \cite{zhao2023survey}. Developing optimization strategies to reduce computational requirements without compromising performance is necessary \cite{bai2022towards}. This may involve new model architectures, data processing methods, and hardware efficiencies, facilitating more scalable LLM deployment.

\textit{Research and Development Directions}
Engaging in collaborations with academic institutions, continuous model training, and adopting user-centric design philosophies are crucial for the ongoing enhancement of LLMs. These efforts are aimed at ensuring models meet the diverse and changing needs of their users.

\noindent \paragraph{Mapping FAIR principles to Data Management Challenges in LLMs} The challenges in managing data for LLMs are complex, yet the FAIR data principles offer targeted solutions that can alleviate some of these issues. For instance, \textit{Findability} enhances the process of identifying pertinent data within vast datasets through the use of enriched metadata and persistent identifiers, streamlining data retrieval and analysis. \textit{Interoperability} facilitates the seamless integration of varying data formats and systems, important for LLM training. This principle also aids in developing algorithms that are more contextually aware by allowing for the combination of different data types and sources, thereby leading to more nuanced and accurate model outputs.

\textit{Accessibility} ensures data is readily available and access is controlled appropriately, allowing for the responsible sharing of data. This principle shows the fine line between promoting open access to data to foster innovation and collaboration, and the necessity to safeguard sensitive information and intellectual property. \textit{Reusability} amplifies the longevity and utility of data beyond its original application. By adhering to legal and regulatory standards, reusability ensures that data remains applicable, reliable, and ethically sound for future endeavors. A detailed mapping of these data management challenges in LLMs to the FAIR data principles is provided in Appendix \ref{App:C}.

\section{Framework for FAIR Data Principles Integration in LLM Development}
\label{proposed}

\paragraph{Rationale for Integrating FAIR Data Principles in LLM Development}

\begin{figure}[ht]
    \centering
    \includegraphics[width=0.95\linewidth]{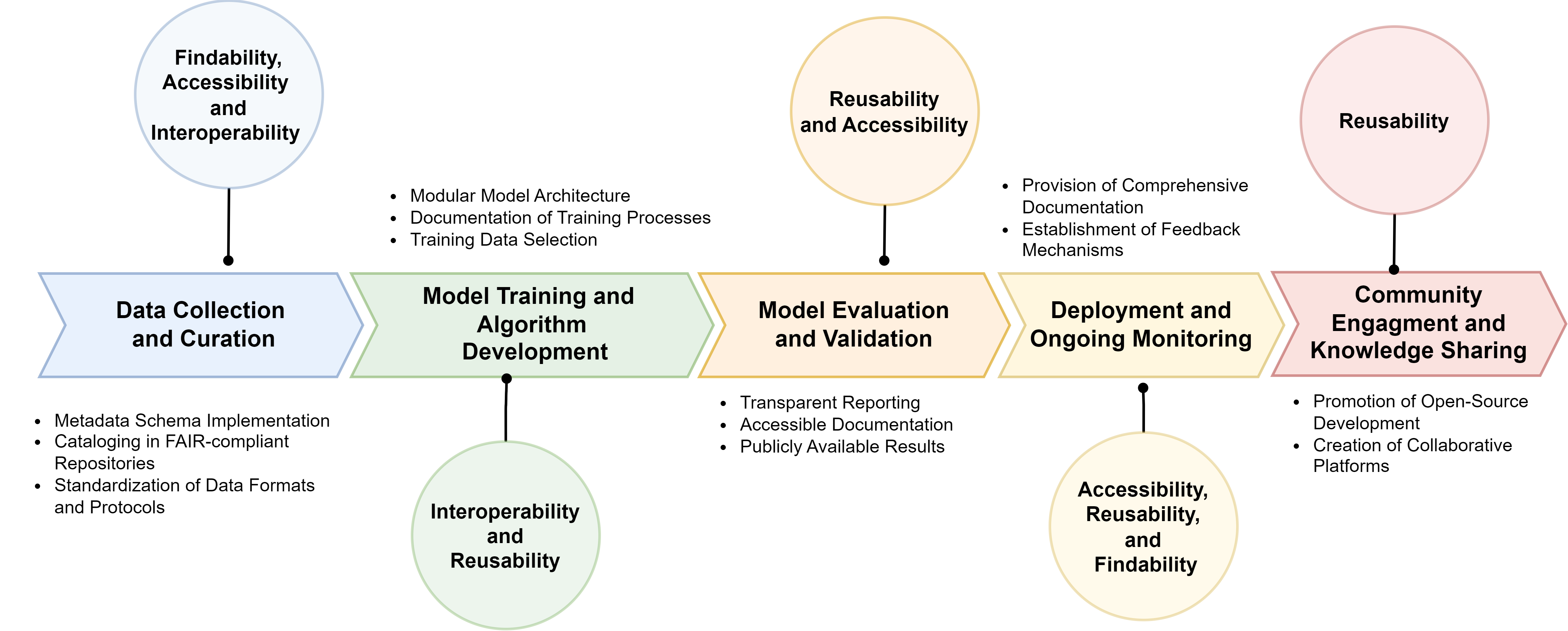}
    \caption{FAIR principles integrated into the LLM lifecycle}
    \label{fig:llm-lifecycle}
\end{figure}

Our research is dedicated to enhancing the training of LLMs by embedding FAIR data principles directly into their training methodologies. The main objective is to develop training datasets for LLMs that are designed to comply with FAIR principles. To realize this, we have devised a framework that interweaves FAIR principles throughout the entire model development lifecycle of LLMs, as depicted in Figure \ref{fig:llm-lifecycle}.  In the following sections, we will elaborate on each phase of this proposed LLM lifecycle, connecting each step specifically to our case study which concentrates on the creation of a FAIR-compliant dataset.

\paragraph{Case Study: Developing a FAIR-Compliant Dataset}
To address the risk of LLMs perpetuating societal biases, our case study focuses on developing a dataset with a conscientious design that proactively identifies biases within the data prior to the training of these models. Adopting FAIR principles in preparing data for LLMs training is essential, though not a full ``FAIRification" of LLMs. This step is crucial for enhancing data quality and model performance, paving the way towards more efficient and ethical AI development. The dataset and model card are available here \footnote{https://huggingface.co/collections/newsmediabias/biasscan-659d681ed7a5bc9d98cde11b} . 

\begin{figure}[ht]
    \centering
    \includegraphics[width=0.9\linewidth]{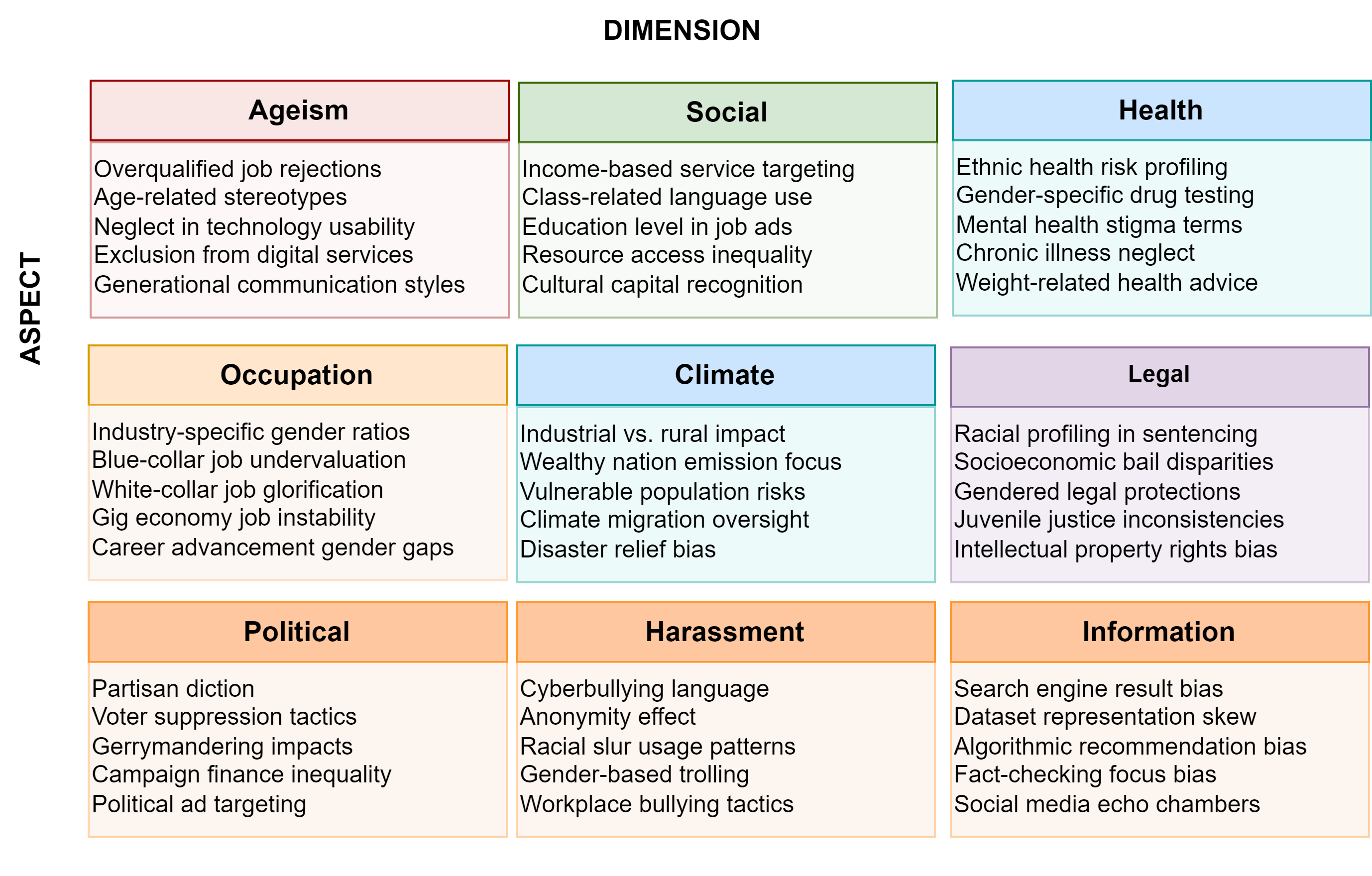}
    \caption{Biases across Multiple Dimensions Explored in this Study}
    \label{fig:biases_dimensions}
\end{figure}

This case study identifies ``bias" in an NLP dataset as a linguistic tendency leading to the unfair representation of specific groups, manifesting as linguistic bias, stereotypes, toxicity, or misinformation \cite{ntoutsi2020bias, raza2024nbias, StereoSet, barikeri2021redditbias, raza2022fake}. Such biases often stem from data collection, processing, and usage methods, potentially causing LLMs to replicate or amplify these biases \cite{MeasuringSocialBiases}. Our analysis covers various bias dimensions, illustrated in Figure \ref{fig:biases_dimensions}, encompassing ageism, occupational jargon, and political rhetoric, among others. This dataset exemplifies how to construct diverse training datasets for LLMs in adherence to FAIR principles.

\noindent \subparagraph{Data Collection and Curation for LLM Development} 
\label{subsec:data_collection_curation}
 \textit{FAIR Principles Achieved:} Findability, Accessibility, Interoperability, and Reusability.

\noindent In line with the FAIR data principles, our dataset is  structured to enhance its \textit{Findability,} featuring comprehensive metadata to facilitate easy discovery. This dataset, sourced from a variety of channels such as news feeds and websites between January and May 2023, encompasses over 50,000 filtered entries. For data curation, we utilized various feeds and hashtags, including \#MediaBias, \#SocialJustice, \#GenderEquality, \#RacialInjustice, \#CulturalDiversity, \#AgeismAwareness, \#ReligiousTolerance, and \#EconomicDisparity, to ensure a wide representation of social issues. This diversity of news articles not only broadens accessibility but also enriches the dataset relevance to current social discourses.

The detailed metadata includes crucial information such as dataset title, description, authors, date of creation, version, and keywords that reflect the dataset scope and purpose. These keywords (e.g., LLMs, Training, Biases, News Media, NLP) are strategically chosen to enhance the dataset retrieval efficiency across various research portals. Additionally, the metadata specifies the data type, whether textual, numerical, or otherwise, further aligning with the \textit{Interoperability} principle by facilitating data integration across different systems.

\begin{figure}[ht]

       \includegraphics[width=.9\linewidth]{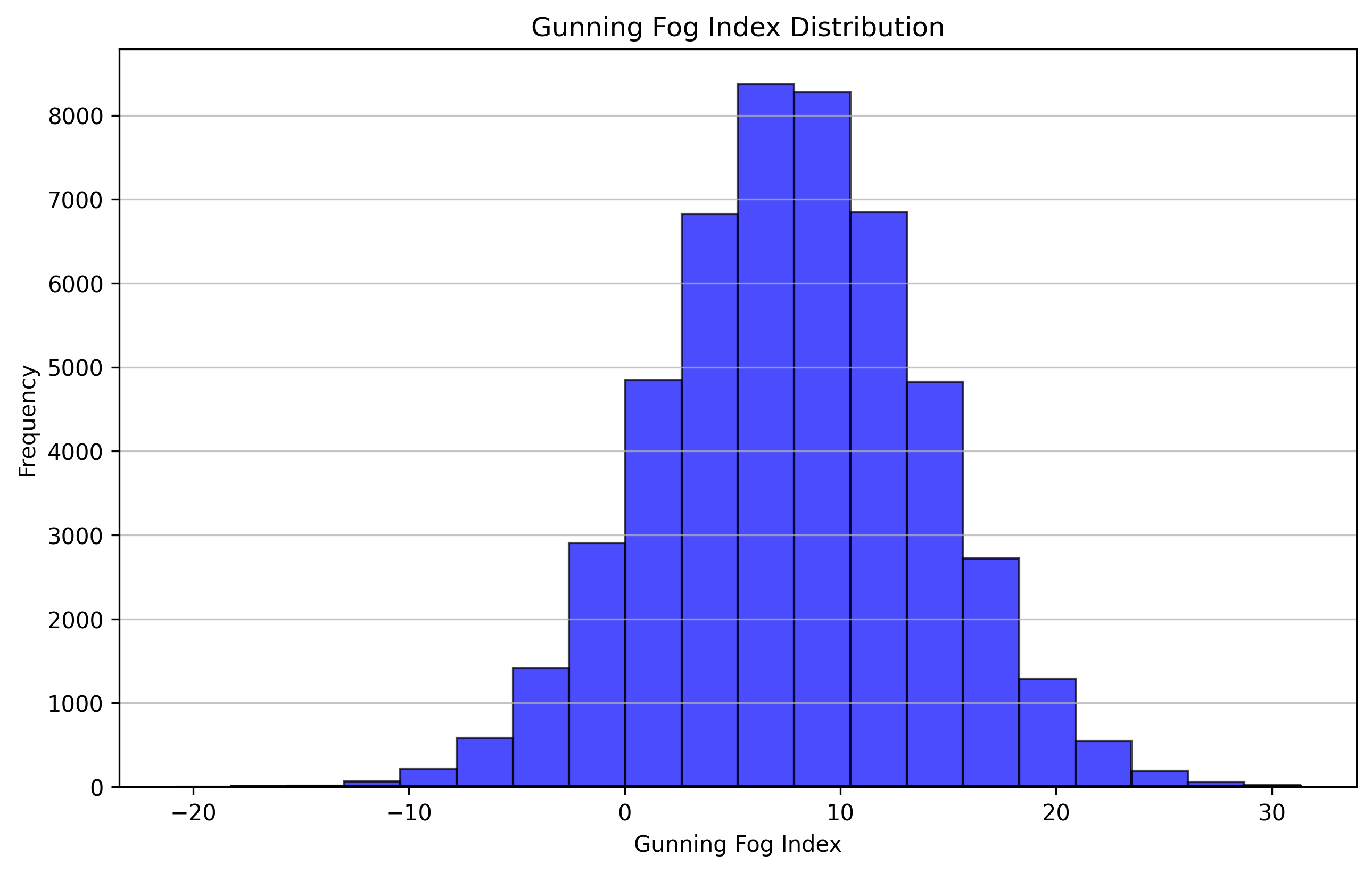}
    \caption{Histogram of the Gunning Fog Index on FAIR-Complaint Dataset. The x-axis denotes the Gunning Fog Index scores, reflecting text complexity, and the y-axis represents the number of samples with each score.}
    \label{fig:gunning_fog_histogram}
    \vspace{5mm} 

    \begin{minipage}{\textwidth} 
    \centering
   
    \begin{tabular}{@{}lp{3cm}p{4.5cm}@{}}
        \toprule
        \textbf{Aspect} & \textbf{Automated Analysis} & \textbf{Expert Review} \\
        \midrule
        Accuracy & 96.5\% & Contextually accurate \\
        Completeness & 93.0\% & High with occasional updates \\
        Consistency & 98.0\% & Uniform across samples \\
        \bottomrule
    \end{tabular}
    \label{tab:data_quality_summary}
     \caption{Data Quality Analysis Summary}
\end{minipage}
\end{figure}

This readability test is crucial for determining the level of education required to comprehend our dataset texts. As illustrated in Figure \ref{fig:gunning_fog_histogram}, the distribution of Gunning Fog Index scores in our dataset shows a normal distribution with a mean score of 7.79. This indicates that the majority of our texts are suitable for readers with at least an eighth-grade education level. Such readability analysis is instrumental in aligning our dataset with the FAIR principle of \textit{Accessibility,} ensuring that the 

\textbf{Annotations} The process of annotating labels and debiasing text initially utilized GPT-3.5 for an automated preliminary screening, efficiently identifying potential bias, toxicity, stereotyping, and harm . Recent works \cite{gilardi2023chatgpt} indicate that AI can excel in labeling tasks, outperforming traditional crowd-sourced methods. In our case, we utilize an extra layer of human review to have quality with speed. This facilitated the early detection of content requiring human review. Following this, a team of 15 experts and students across various disciplines crafted comprehensive annotation guidelines emphasizing accuracy and sensitivity. These guidelines assist annotators in conducting a thorough review to mitigate any biases, ensuring content fairness and respectfulness.

Annotators engage in a nuanced review process, identifying and revising instances of bias, toxicity, stereotyping, and harm based on a wide range of attributes, guided by principles of neutrality, respect, and inclusivity. This detailed annotation effort is supported by a commitment to ongoing feedback, expert reviews, and inter-annotator agreement (IAA) checks, ensuring high-quality annotations evidenced by a Cohen's Kappa score above 0.75. Examples of IAA agreements across main bias dimensions are provided in Appendix \ref{iaa}, illustrating the robustness of our annotation methodology.

The robustness of our dataset is established through a dual-stage quality assessment: automated analysis for initial screening of 10,000 statistically significant entries, and expert review of 500 stratified random samples for deeper insights. This method ensures the dataset accuracy, completeness, and consistency, detailed in Figure \ref{tab:data_quality_summary}.

For \textit{Interoperability}, the data is formatted for seamless integration for various ML tasks, such as binary and multi-label classifiers, question answering (QA) system and debiased language generation task (debiasing), as detailed in Table \ref{tab:dataset-formats}. This enables researchers to use this data in diverse analytical contexts, facilitating cross-domain research and development. The dataset schema and examples on dataset formats are given in Appendix \ref{App:A}.

\begin{table}[h]
\centering
\small
\caption{Specialized Data Formats for Interoperability}
\label{tab:dataset-formats}
\begin{tabular}{>{\bfseries}l l}
\toprule
\textbf{Data Format} & \textbf{Model Type} \\
\midrule
Classification Format & Binary Classifier \\
CoNLL Format & Multi-label Token Classification\\
SQuAD Format & Question Answering System \\
Counterfactual Formatting & Debiasing Model \\
Sentiment Analysis Format & Sentiment Classifier \\
Toxicity Classification Format & Toxicity Classifier \\
\bottomrule
\end{tabular}
\end{table}

Furthermore, we have stored our dataset in repositories such as Huggingface, Zenodo, and Figshare, which not only adhere to metadata standards but also ensure its long-term \textit{Accessibility}. Overall, our data collection and curation strategy aligns with the FAIR principles and enhance the practical utility of our dataset for LLM development.

\noindent\subparagraph{Model Training and Algorithm Development}
\label{sec:model_training_and_algo_dev}

 \textit{FAIR Principles Achieved:} Interoperability and Reusability.

\noindent The model training and algorithm development phase is crucial for adhering to FAIR principles, particularly interoperability and reusability. We strategically deploy various LLMs, fine-tuned on 10,000 statically significant data sample from our main data, for tasks ranging from sentiment analysis and QA to debiasing (language generation), and develop them with a modular design for enhanced \textit{Reusability} across projects.

\textit{Interoperability} is achieved by using common frameworks like TensorFlow and PyTorch and standardizing data formats for inputs and outputs. Every task performed by LLMs, such as classification and debiasing, is executed following the preparation of specific training formats tailored to each task. This ensures seamless integration of our models with various systems and datasets. We further enhance this with the inclusion of model cards for each LLM, providing essential information such as model purpose, performance, and usage guidelines, thus aiding in understanding and adoption.

We prepare API documentation that provides details for every aspect of training and development, including parameter settings and algorithm modifications. This transparency facilitates scientific validation and progress \cite{jacobsen2020fair}. We also foster community collaboration by sharing our findings, models, and tools. 

\noindent\subparagraph{Model Evaluation and Validation}
 \textit{FAIR Principles Achieved:}  Reusability and Accessibility.

\begin{figure}[h]
\centering
\includegraphics[width=0.8\textwidth]{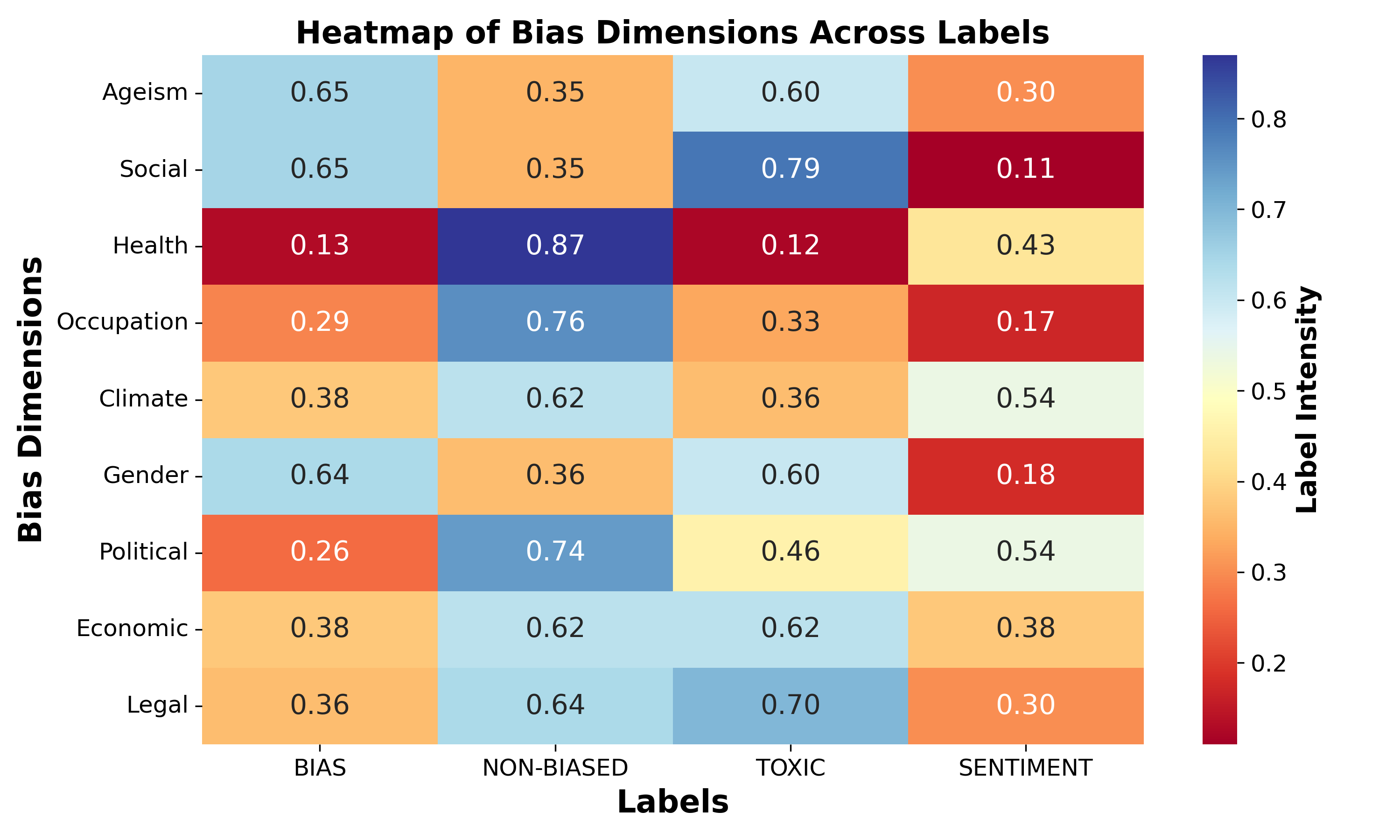}
\caption{Heatmap Visualization: the prevalence and intensity of different types of biases, such as ageism, gender, and political, across various classifications like bias, non-biased, toxic, and sentiment within a dataset.}
\label{fig:heatmap}
\end{figure}

\noindent In our evaluation and validation phase, we prioritize the FAIR principles of \textit{Reusability} and \textit{Accessibility} to ensure that our models and findings can be widely utilized and understood. Transparency in reporting and the provision of accessible documentation are key aspects of our methodology. All results, including bias analysis and model benchmarking, are made publicly available, allowing for comprehensive observation and application by the wider research community. An example of this approach is presented in our detailed bias analysis (Figure \ref{fig:heatmap}) and the benchmarking table (Table \ref{tab:combined_model_debiasing}), which illustrate our commitment to these principles.

 In Table \ref{tab:combined_model_debiasing}, we present performance metrics across various tasks, demonstrating the effectiveness of our classifiers in toxicity detection, bias classification, sentiment analysis, multi-label token classification, and QA capabilities. Additionally, we integrated a debiasing process using LLaMa 2B Chat \cite{touvron2023llama}, re-evaluating our models on the same test set post-debiasing. This allowed us to assess the impact of debiasing on reducing toxicity and bias, and its influence on sentiment analysis. The debiasing process notably improved scores and indicated the efficacy of the debiasing intervention. 
 
\begin{table}[h]
\setlength{\tabcolsep}{4pt} 
\centering
\small
\caption{Comprehensive Analysis of Language Model Performance and Impact of Debiasing: the first section details performance metrics of various models on different tasks, fine-tuned for optimal settings. The second section highlights the effect of debiasing on toxicity, bias classification, and sentiment scores.}
\label{tab:combined_model_debiasing}
\renewcommand{\arraystretch}{1} 
\begin{tabular}{@{}lccc@{}}
\toprule
\multicolumn{4}{c}{\textbf{Performance Metrics of Language Models. Higher the score, better the performance.}} \\
\textbf{Model (Type)} & \textbf{Accuracy} & \textbf{Precision} & \textbf{F1-Score} \\ \midrule
Binary Bias Classifier (BERT-large-uncased) & 92\% & 89\% & 90\% \\
NER Model (RoBERTa-large-uncased) & 95\% & 93\% & 94\% \\
QA System (BERT-base-uncased) & 88\% & 87\% & 86\% \\
Sentiment Classifier (BERT-large-uncased) & 90\% & 91\% & 90\% \\
Toxicity Classifier (BERT-large-uncased) & 91\% & 90\% & 91\% \\
LLaMa2 (7b-chat-hf)& 93\%& 94\%&93\%\\ 
\midrule
\multicolumn{4}{c}{\textbf{Impact of Debiasing. Lower scores post-debiasing is better.}} \\
\textbf{Metric} & \textbf{Pre-Debiasing Score} & \textbf{Post-Debiasing Score} & \\ \midrule
Toxicity Score & 75\% & 24\%& \\
Bias Classification Score & 80\% & 55\% & \\
Sentiment Score & -0.3 & 0.2 & \\
\bottomrule
\\
\end{tabular}

\justifying
The results in the table demonstrate that all models initially exhibit strong classification performance. However, after the debiasing process, a noticeable decline in classification performances is observed across these models when applied to the debiased dataset. This suggests that the original training was heavily reliant on biased data, and the introduction of debiased data disrupts the model's classification ability, leading to a reduction in accuracy. This result aligns very much with the recent safety benchmarking in recent works \cite{touvron2023llama}, which shows there is a trade-off highlighting the impact of debiasing on model performance.
\end{table}

Overall, these  models' evaluation and validation phase, emphasize transparent reporting and public availability of models and the detailed results, aligns with the FAIR principles of Reusability and Accessibility.

\noindent\subparagraph{Deployment and Ongoing Monitoring} 
 \textit{FAIR Principles Achieved:} Accessibility, Reusability, and Findability.

\noindent In the post-deployment phase, we emphasize the FAIR principles of\textit{ Accessibility, Reusability, }and \textit{Findability}. Our approach includes providing comprehensive documentation and clear usage guidelines to enhance the model findability and \textit{Accessibility}. We employ version control systems to ensure easy tracking and retrieval of the latest model versions, supporting both findability and \textit{Reusability.}

To facilitate user engagement and model effectiveness, we offer extensive training materials and support, making the models more accessible to a diverse user base. Regular reporting of performance metrics offers insights into the models' ongoing effectiveness and areas for improvement, aiding in their reusability. We also provide integration support with various tools and platforms, ensuring that our models can be easily adopted in different technological contexts. These practices collectively ensure that our deployed models adhere to the FAIR principles, maintaining their utility and efficacy in a dynamic, user-centric environment.

\noindent\subparagraph{Community Engagement and Collaborative Development} 
 \textit{FAIR Principles Achieved:}  Findability, Accessibility, Interoperability, and Reusability.

\noindent In our efforts to foster community engagement and collaborative development, we align with the FAIR principles effectively.  We provide open-source development to facilitate collaborative innovation. Our datasets are provided in accessible formats such as JSON and CSV and are indexed with rich metadata. This approach ensures that our LLMs are developed in environments conducive to collaborative progress. The datasets are disseminated under the \textbf{\textit{Creative Commons Attribution-NonCommercial 4.0 International (CC BY-NC 4.0)}} license \cite{creative_commons}, aligning our work with the FAIR principles of data stewardship.

\noindent \subparagraph{Upholding FAIR Principles}
We provide a comprehensive analysis in Table \ref{table:fair_compliance_llms_models}, showcasing our commitment to upholding the FAIR data principles to enhance the dataset's usefulness for both present and future research endeavors.

\begin{center}
\small
\begin{longtable}{|m{1.5cm}|m{4.5cm}|m{0.5cm}|m{5cm}|}
\caption{FAIR Compliance in Dataset and LLM Development}\label{table:fair_compliance_llms_models} \\
\hline
\textbf{Principle} & \textbf{FAIR Compliance Criteria} & \textbf{Met} & \textbf{Remarks (Implementation Details)} \\
\hline
\endfirsthead

\multicolumn{4}{c}{{\bfseries Table \thetable\ Continued from previous page}} \\
\hline
\textbf{Principle} & \textbf{FAIR Compliance Criteria} & \textbf{Met} & \textbf{Remarks (Implementation Details)} \\
\hline
\endhead

\hline
\endfoot

\hline
\endlastfoot

Findability & 
   F1: Rich, descriptive metadata \newline
   F2: Standardized data indexing \newline
   F3: Documented data sources \newline
   F4: Persistent identifiers for data and models
& 
   \checkmark &
   F1: Metadata includes descriptions for data and models. \newline
   F2: Data and models indexed with relevant keywords. \newline
   F3: Clear documentation of data sources and model development. \newline
   F4: Unique identifiers for data and models.
\\
\hline
Accessibility & 
   A1: Clearly defined access protocols \newline
   A2: Long-term data and model preservation \newline
   A3: Data and models available in accessible formats
 & 
   \checkmark &
   A1: Open-access repository for data and models. \newline
   A2: Long-term preservation of data and models. \newline
   A3: Data in CSV, JSON, XML; Models accessible via APIs.
\\
\hline
Interoper-ability & 
   I1: Use of standard data and model formats \newline
   I2: Adherence to data and model exchange standards
 & 
   \checkmark &
   I1: Data in standard formats; Models compatible with common ML frameworks. \newline
   I2: Adheres to standard data and model exchange protocols.
\\
\hline
Reusability & 
   R1: Detailed metadata for data and models \newline
   R2: Adherence to ethical and transparency standards \newline
   R3: Licensing information for data and model usage
 & 
   \checkmark &
   R1: Metadata includes usage guidelines for data and models. \newline
   R2: Focus on ethical data use and transparent model training. \newline
   R3: Data licensed under CC BY-NC 4.0; Models with open-source licenses.
\\
\hline
\end{longtable}
\end{center}

\section{Discussion}

\paragraph{Limitations of the FAIR Data Principles in Addressing Data Challenges}
The FAIR data principles is a significant step towards enhancing the openness and efficiency of data usage in research and related fields. However, their effectiveness is constrained by several factors. Primarily, while the FAIR principles improve data accessibility and usability, they do not automatically guarantee data quality or validity. The absence of mechanisms for ensuring accuracy, completeness, or reliability could lead to the spread of low-quality data, negatively impacting research and decision-making. Additionally, FAIR principles, though encouraging data sharing, may not adequately address the ethical and privacy concerns associated with sensitive data like health records or personal information, which could result in privacy violations.

Furthermore, the implementation of FAIR principles requires considerable resources, including necessary infrastructure and expertise. This poses a significant challenge, especially for smaller institutions or those in developing countries, potentially exacerbating the digital divide. The broad application of FAIR principles may not be suitable for all scientific disciplines, as different fields might require tailored data management approaches not fully encompassed by these general guidelines. The effectiveness of these principles also depends heavily on the awareness and training of data handlers, where a lack of such training can be a major barrier to adoption. Moreover, the absence of universally accepted standards for evaluating 'FAIRness' leads to an inconsistent application of these principles.

The diversity of data formats creates technical challenges in achieving seamless interoperability, where the emphasis on data sharing might overshadow other vital aspects of data stewardship, such as privacy considerations. Finally, the principles can conflict with intellectual property rights and commercial interests. The concept of free data access and reuse might challenge proprietary research, leading to resistance from certain sectors. This necessitates compliance with various legal and regulatory frameworks, adding another layer of complexity to the application of FAIR principles.

\paragraph{Strategies for Mitigating Limitations and Enhancing Data Utility} 
In addressing the limitations of FAIR principles and enhancing data utility, a multifaceted strategy is essential. This involves refining data quality through rigorous validation processes, ensuring accuracy and reliability. Simultaneously, balancing data accessibility with ethical and privacy considerations is crucial, there is need to employ proper anonymization techniques and strict privacy protocols. Emphasizing data preservation and sustainability alongside sharing, broadens the scope of stewardship. . 

\paragraph{Limitation of the Study and Future Perspectives }
Our Fair-Compliant dataset, designed to identify and mitigate known biases, may inadvertently overlook emerging biases, thereby emphasizing the need for ongoing revision and vigilant monitoring \cite{MeasuringSocialBiases}. Concurrently, scaling the dataset to match the complexity of advanced LLMs, while adhering to ethical standards, emerges as a substantial challenge. This endeavor is compounded by the imperative to mitigate biases in model interpretations, bolster interpretability, and ensure robust data privacy and security \cite{zhao2023explainability, chen2023can}. Therefore, future research directions should consider creating dynamic mechanisms for dataset updates \cite{wilson2023abstract}, advancing bias mitigation techniques, aligning datasets with evolving LLM technologies, exploring scalable dataset maintenance solutions, and broadening the spectrum of LLM applications. Integral to this journey is the development of comprehensive ethical guidelines and the fostering of collaborative research, both of which are pivotal for the responsible evolution of LLM technology.

\section{ Conclusion}
Our study incorporates FAIR data principles into LLM training and development, improving data management and model training. This approach has yielded a versatile dataset that shows the process of integrating FAIR principles into the data building and that can be used for LLM training. While our dataset and case study offer guidance, they do not fully tackle the `FAIRification' of LLMs. Advancements in LLM research must focus on data and the applicability of FAIR principles. This includes updating datasets to capture emerging trends, enhancing bias detection, adapting to novel LLM architectures, improving model interpretability, and formulating ethical guidelines. Our efforts contribute to the responsible advancement of AI, aiming to forge more ethical and efficient AI tools that serve diverse communities.

\section*{Acknowledgements}
\noindent Resources used in preparing this research were provided, in part, by the Province of Ontario, the Government of Canada through CIFAR, and companies sponsoring the Vector Institute.

\bibliographystyle{plainnat}
 \bibliography{references}


\newpage
\appendix

\setcounter{figure}{0}
\setcounter{table}{0}
\section*{Appendices}

\subsection{Acronyms and Full Forms}
\label{App: acronyms}

 \href{https://www.academia.edu/}{Academia.edu}, Academia.edu;
 \href{https://ai-4-all.org/}{AI4ALL}, AI4ALL;
 \href{https://www.algolia.com/}{Algolia}, Algolia;
 \href{https://aws.amazon.com/pm/serv-s3}{Amazon S3}, Amazon Simple Storage Service;
 \href{https://atlas.apache.org}{Apache Atlas}, Apache Atlas;
 \href{https://lucene.apache.org/}{Apache Lucene}, Apache Lucene;
 \href{https://nifi.apache.org/}{Apache NiFi}, Apache NiFi;
 \href{https://solr.apache.org/}{Apache Solr}, Apache Solr;
 \href{https://www.archivematica.org/en/}{Archivematica}, Archivematica;
 \href{https://ckan.org/}{CKAN}, Comprehensive Knowledge Archive Network;
 \href{https://clockss.org/}{Clockss}, Controlled LOCKSS;
 \href{https://www.collibra.com/us/en}{Collibra}, Collibra;
 \href{https://creativecommons.org/}{Creative Commons}, Creative Commons;
 \href{https://www.crossref.org/}{Crossref}, Crossref;
 \href{https://www.dspace.com/en/inc/home.cfm}{DSpace}, DSpace;
 \href{https://www.dublincore.org/}{Dublin Core}, Dublin Core Metadata Initiative;
 \href{https://projects.iq.harvard.edu/provenance-at-harvard/tools}{Data Provenance Tools}, Data Provenance Tools;
 \href{https://datacite.org/}{DataCite}, DataCite;
 \href{https://powerplatform.microsoft.com/en-ca/dataverse/}{Dataverse}, Microsoft Dataverse;
 \href{https://www.eprints.org/uk/}{EPrints}, EPrints;
 \href{https://eml.ecoinformatics.org/}{Ecoinformatics}, Ecoinformatics;
 \href{https://www.elastic.co/}{Elasticsearch}, Elasticsearch;
 \href{https://www.fged.org/projects/miame}{FGED}, Functional Genomics Data Society;
 \href{https://figshare.com/}{Figshare}, Figshare;
 \href{https://gdpr-info.eu/}{GDPR}, General Data Protection Regulation;
 \href{https://one.google.com/}{Google Cloud Storage}, Google Cloud Storage;
 \href{https://graphql.org/}{GraphQL}, GraphQL;
 \href{https://www.hl7.org/fhir/}{HL7 FHIR}, Health Level Seven Fast Healthcare Interoperability Resources;
 \href{https://www.iedb.org/}{IEDB}, Immune Epitope Database;
 \href{https://en.wikipedia.org/wiki/ISO/IEC_27001}{ISO/IEC}, International Organization for Standardization/International Electrotechnical Commission;
 \href{https://www.lockss.org/}{LOCKSS}, Lots of Copies Keep Stuff Safe;
 \href{https://www.ncbi.nlm.nih.gov/}{NCBI}, National Center for Biotechnology Information;
 \href{https://www.openarchives.org/pmh/}{OAI-PMH}, Open Archives Initiative Protocol for Metadata Harvesting;
 \href{https://omeka.org/}{Omeka}, Omeka;
 \href{https://www.onetrust.com/}{OneTrust}, OneTrust;
 \href{https://openai.com/policies/supplier-code}{OpenAI Ethics Guidelines}, OpenAI Ethics Guidelines;
 \href{https://www.openapis.org/}{OpenAPI}, OpenAPI Initiative;
 \href{https://openrefine.org/}{OpenRefine}, OpenRefine;
 \href{https://orcid.org/}{ORCID}, Open Researcher and Contributor ID;
 \href{https://www.owl.co/}{OWL}, Web Ontology Language;
 \href{https://www.portico.org/}{Portico}, Portico;
 \href{https://www.w3.org/TR/prov-dm/}{PROV-DM}, Provenance Data Model;
 \href{https://www.w3.org/RDF/}{RDF}, Resource Description Framework;
 \href{https://www.re3data.org/}{RE3data}, Registry of Research Data Repositories;
 \href{https://www.researchgate.net/}{ResearchGate}, ResearchGate;
 \href{https://www.responsible.ai/}{Responsible AI}, Responsible AI;
 \href{https://www.ibm.com/topics/rest-apis}{REST}, Representational State Transfer;
 \href{https://schema.org/}{schema.org}, Schema.org;
 \href{https://www.guru99.com/soap-simple-object-access-protocol.html}{SOAP}, Simple Object Access Protocol;
 \href{https://www.w3.org/TR/sparql11-query/}{SPARQL}, SPARQL Protocol and RDF Query Language;
 \href{https://www.talend.com/}{Talend}, Talend;
 \href{https://trustarc.com/}{TrustArc}, TrustArc;
 \href{https://xod.io/}{XOD}, eXtensible ontology development;
 \href{https://www.w3schools.com/xml/xsl_intro.asp}{XSLT}, Extensible Stylesheet Language Transformations;
 \href{https://zenodo.org/}{Zenodo}, Zenodo;
 \href{https://grpc.io/}{gRPC}, gRPC Remote Procedure Calls.

\subsection{Data Management Challenges in LLMs and Corresponding FAIR Principles}
\label{App:C}
\begin{table}[h]
\centering
\caption{Data Management Challenges in LLMs and Corresponding FAIR Principles}
\label{tab:fair_mapping}
\begin{tabular}{|m{0.3\linewidth}|m{0.7\linewidth}|}
\hline
\textbf{Data Management Challenge} & \textbf{FAIR Principle Addressed} \\
\hline
Vast and Complex Datasets & Findability through detailed metadata and persistent identifiers to enhance data discovery . \\
\hline
Data Quality and Bias & Accessibility with ethical access protocols to provide high-quality, unbiased data. \\
\hline
Privacy and Ethical Concerns & Reusability with clear legal and ethical documentation to uphold privacy and ethics. \\
\hline
Data Annotation and Labeling & Interoperability through standardized data formats for reliable annotation and labeling. \\
\hline
Data Accessibility and Sharing & Accessibility to ensure a balance between open data sharing and protection of proprietary information . \\
\hline
Legal and Regulatory Compliance & Reusability to align data management with legal standards for future use . \\
\hline
Contextual Awareness & Interoperability for enhanced NLP algorithms to accurately capture and interpret context. \\
\hline
Accuracy and Reliability & Accessibility and Reusability to ensure mechanisms for improved fact-checking and consistent data quality. \\
\hline
Ethical and Fair Use & Accessibility with advanced bias detection algorithms to promote fairness . \\
\hline
Interactivity and Personalization & Findability and Accessibility for adaptive learning from user interactions for personalized experiences. \\
\hline
Language and Cultural Sensitivity & Interoperability to support expanded language models and cultural datasets for inclusivity . \\
\hline
Technical and Scalability & Reusability for efficient processing strategies that facilitate the scalable use of computational resources. \\
\hline
\end{tabular}
\end{table}

\subsection{Dataset Schema and Formats}
\label{App:A}
The dataset has multiple attributes like\textit{ Text, Dimension, Biased Words, Aspect, Bias Label, Sentiment, Toxic, Identity Mention, Debasied Text}.

  \noindent "Text": "Lawyers are always manipulative and cannot be trusted.", \\
    "Dimension": "Professional Integrity", \\
    "Biased Words": "always, manipulative, cannot be trusted",\\
    "Aspect": "Trustworthiness of lawyers",\\
    "Bias Label": "BIASED",\\
    "Sentiment": "Negative",\\
    "Toxic": "Yes",\\
    "Identity Mention": "Lawyers",\\
    "Debiased Text": "Trustworthiness is an individual trait  and varies among professionals, including lawyers."

We provide examples for each data format, demonstrating the use of dataset attributes in diverse machine learning models:

\begin{table}[h]
\centering
\small
\caption{Examples of Dataset Formats}
\label{tab:dataset-format-examples}
\resizebox{\textwidth}{!}{%
\begin{tabular}{>{\bfseries}l p{10cm}}
\toprule
Format & Example \\
\midrule
Classification & \{ "Text": "Politicians are often corrupt.", "Bias Label": "BIASED", "Sentiment": "Negative" \} \\
CoNLL & \{ "Text": "Doctors are always caring.", "Biased Words": ["always", "caring"], "Identity Mention": "Doctors" \} \\
SQuAD & \{ "Text": "Why are lawyers untrustworthy?", "Aspect": "Trustworthiness", "Debiased Text": "Trustworthiness varies individually." \} \\
Counterfactual & \{ "Text": "Young people are irresponsible with money.", "Biased Words": ["irresponsible"], "Debiased Text": "Financial habits vary by individual." \} \\
Sentiment/Toxicity & \{ "Text": "Politicians are dishonest.", "Sentiment": "Negative", "Toxic": "Yes" \} \\
\bottomrule
\end{tabular}
}
\end{table}

\subsection{Inter Annotators Agreement}
\label{iaa}
\begin{figure}[h]
\centering
\includegraphics[width=0.75\textwidth]{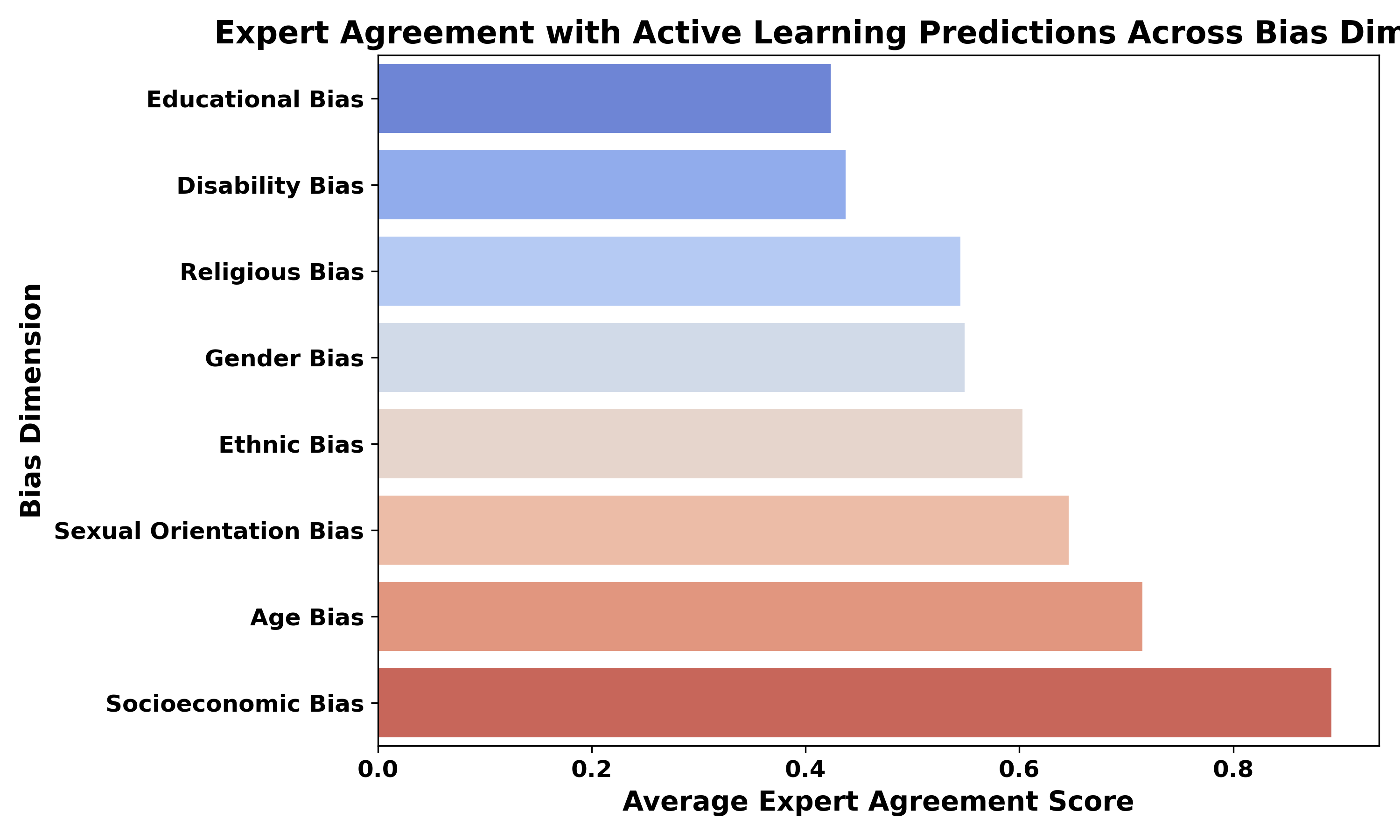}
\caption{Expert Agreement Across Bias Dimensions. The bar graph quantifies the concordance between domain experts evaluations and the model's predictions.}
\label{fig:iaa}
\end{figure}

Figure \ref{fig:iaa} depicts the consensus among experts on different dimensions of bias, with the most significant agreement observed in the area of 'Socioeconomic Bias'. This figure highlights the level of concordance among annotators regarding the various aspects of bias. 

\end{document}